\documentclass[lettersize,journal]{IEEEtran}
\usepackage{amsmath,amsfonts}
\usepackage{algorithmic}
\usepackage{array}
\usepackage[caption=false,font=normalsize,labelfont=sf,textfont=sf]{subfig}
\usepackage{textcomp}
\usepackage{stfloats}
\usepackage{url}
\usepackage{verbatim}
\usepackage{hyperref}
\usepackage{graphicx}
\usepackage{booktabs}
\usepackage[table,xcdraw]{xcolor}
\usepackage{algorithm}
\usepackage{algorithmic}
\usepackage{booktabs} 
\usepackage{amsmath} 
\usepackage{pifont}     
\newcommand{\cmark}{\ding{51}}  
\newcommand{\xmark}{\ding{55}} 
\usepackage{newfloat}
\usepackage{listings}
\usepackage{times}  
\usepackage{helvet}  
\usepackage{courier}  

\usepackage{tabularx, booktabs}

\usepackage{natbib}  
\usepackage{caption} 
\title{UMIGen: A Unified Framework for Egocentric Point Cloud Generation and Cross-Embodiment Robotic Imitation Learning}
\author{Yan Huang, Shoujie Li, Xingting Li, Wenbo Ding
\thanks{\textit{Yan Huang , Shoujie Li contributed equally to this work.} (Corresponding author: Wenbo Ding, ding.wenbo@sz.tsinghua.edu.cn)}

\thanks{Yan Huang, Shoujie Li, Xingting Li and Wenbo Ding are with Shenzhen International Graduate School, Tsinghua University, Shenzhen 518055, China.}}

\def\BibTeX{{\rm B\kern-.05em{\sc i\kern-.025em b}\kern-.08em
    T\kern-.1667em\lower.7ex\hbox{E}\kern-.125emX}}
\usepackage{balance}
\begin{document}

\maketitle

\begin{abstract}
Data-driven robotic learning faces an obvious dilemma: robust policies demand large-scale, high-quality demonstration data, yet collecting such data remains a major challenge owing to high operational costs, dependence on specialized hardware, and the limited spatial generalization capability of current methods.  
The Universal Manipulation Interface~(UMI) relaxes the strict hardware requirements for data collection, but it is restricted to capturing only RGB images of a scene and omits the 3D geometric information on which many tasks rely.  
Inspired by DemoGen, we propose \textbf{UMIGen}, a unified framework that consists of two key components: (1) Cloud-UMI, a handheld data collection device that requires no visual SLAM and simultaneously records point cloud observation–action pairs; (2) a visibility-aware optimization mechanism that extends the DemoGen pipeline to egocentric 3D observations by generating only points within the camera’s field of view. These two components enable efficient data generation that aligns with real egocentric observations and can be directly transferred across different robot embodiments without any post-processing. Experiments in both simulated and real-world settings demonstrate that UMIGen supports strong cross-embodiment generalization and accelerates data collection in diverse manipulation tasks.

\end{abstract}


\begin{figure*}[h]
	\centering
	\includegraphics[width=1\linewidth]{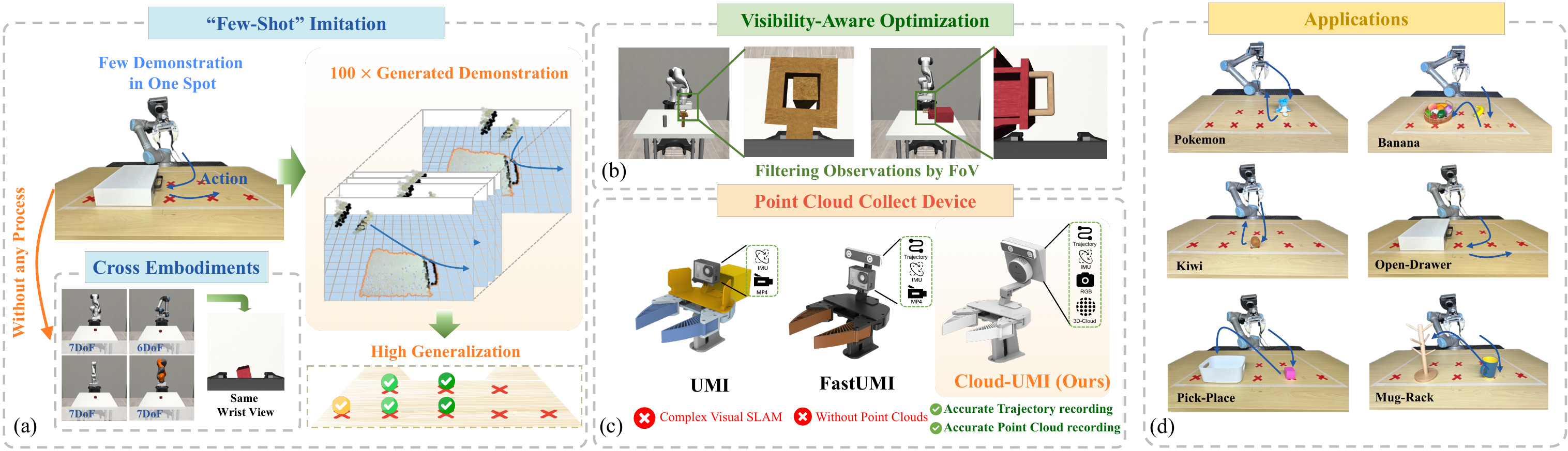}
	\caption{ Overview of UMIGen. (a) From a few wrist-view demonstrations, UMIGen generates diverse samples that generalize spatially and support transfer across robot embodiments sharing the same wrist viewpoint. 
    (b)  During augmentation, only the points within the camera’s field of view are kept. This makes the generated observations realistic and consistent with what the wrist-mounted camera can actually see.
    (c) Cloud-UMI, a low-cost handheld data collection device that fuses a depth sensor with a tracking module, eliminating the need for complex visual-SLAM or external motion capture systems.
    (d) Experiments and applications using UMIGen. Curved arrows trace the end-effector trajectory.}
    \label{fig:UMIGen_overview} \label{head}
\end{figure*}

\section{Introduction}

Training visuomotor policies from images or point clouds has delivered impressive manipulation results in recent years~\cite{chi2023diffusion,ze20243d,fu2024mobile}. Such policies, however, remain highly data hungry. Robust deployment demands hundreds of human demonstrations that span diverse object configurations and environmental conditions~\cite{o2024open,lin2024data} in complex tsaks. The burden becomes even heavier for long‐horizon or multi‐task settings, where real systems may require several thousand trials for a single skill~\cite{zhao2024aloha}. Progress in robot learning is often throttled by the cost of collecting sufficiently diverse data.

3D perception offers a promising path toward stronger generalisation. Point clouds and other 3D observations provide explicit geometry that captures object shape, pose, and spatial relations. Recent frameworks such as PerAct~\cite{shridhar2023perceiver}, GNFactor~\cite{ze2023gnfactor}, and 3D Diffuser Actor~\cite{ke20243d} verify these advantages, while foundation models like FP3 further consolidate the momentum behind point cloud representations~\cite{yang2025fp3}. Yet most real‐world 3D datasets remain narrow in scope. Droid, for instance, gathers large demonstrations with depth sensing, but its fixed arms and carefully calibrated rigs hinder scale and cross‐platform deployment~\cite{khazatsky2025droidlargescaleinthewildrobot}. The field therefore lacks a universally applicable solution for egocentric 3D collection that is robot agnostic, cost effective, and compatible with modern point cloud policies.

Portable pipelines such as the Universal Manipulation Interface (UMI) and FastUMI attempt to lower this barrier by allowing operators to tele‐demonstrate with a handheld RGB camera~\cite{chi2024universal,wu2024fast}. Although these systems reduce hardware complexity, they sacrifice geometric richness; image observations alone cannot recover the spatial structure now known to be critical. The community thus faces an information gap: acquiring large‐scale egocentric point clouds still requires expensive robotics, sophisticated simultaneous localisation and mapping~\cite{Campos_2021}, visual–inertial odometry systems~\cite{chi2024universal}, or labour‐intensive calibration~\cite{fu2024mobile}.

Data augmentation emerges as an attractive remedy. DemoGen shows that spatially transforming a small set of trajectories can synthesize extensive task variations~\cite{xue2025demogen}. Unfortunately, its full‐scene and static‐viewpoint assumptions break down in wrist‐mounted egocentric settings, where visibility is partial and constantly changing. These observations raise two concrete questions. First, how can we capture egocentric point cloud demonstrations without resorting to fixed robotic platforms. Second, how can we expand such partial‐view data into task-diverse trajectories.


We address both questions with UMIGen, a unified framework for efficient egocentric point cloud generation and imitation learning. UMIGen combines a low-cost handheld device with an efficient data generation pipeline. The device integrates an Intel RealSense L515 depth sensor and a T265 tracking camera, and supports point clouds collection in both the camera or robot base coordinates. For data generation, we extend DemoGen with visibility-aware optimization (VAO) mechanism. It discards augmented points that fall outside the wrist camera’s field of view (FoV), resulting in point clouds that better match real egocentric observations. Our contribution are mainly three parts:

\begin{itemize}
    \item A universal handheld data collection device that records paired point-cloud observations and actions without relying on robot hardware or visual SLAM.
    
    \item Introduce VAO mechanism that enhances the demonstration generation pipeline by leveraging the wrist camera's actual FoV to generate egocentric point cloud observations that are naturally aligned with what the robot perceives in the real world.
    
    \item Comprehensive experiments in simulation and on physical robots that benchmark cross-embodiment generalisation, validate data generation efficiency, and demonstrate scalable synthesis across diverse tasks.  
\end{itemize}

\section{Related Work}

\subsection*{a) Robotic Data Collection System}



An intuitive approach to data collection in robotics is teleoperation~\cite{mandlekar2018roboturk}, where a human demonstrator directly controls the robot to perform tasks, thereby generating corresponding demonstration data. Teleoperation can be implemented through a variety of modalities and control interfaces, such as master-slave arm configurations~\cite{zhao2023learningfinegrainedbimanualmanipulation,wu2024gello,fu2024mobile} and kinesthetic teaching. 

In recent years, alternative teleoperation interfaces including SpaceMouse devices~\cite{chi2023diffusion,zhu2023viola}, augmented or virtual reality (AR/VR) control setups~\cite{seo2023deep,rosete2023latent}. Some systems also incorporate multimodal feedback mechanisms such as haptic~\cite{toedtheide2023force} cues or force feedback~\cite{liu2025factrforceattendingcurriculumtraining} to facilitate smoother and more informative data collection across diverse manipulation scenarios.

Recently, handheld data collection devices have enabled more convenient in-the-wild data collection~\cite{chi2024universal}. Several studies have augmented these handheld systems with additional sensors, such as tactile~\cite{zhu2025touchwildlearningfinegrained} or depth sensors~\cite{wu2024fast}, to capture multimodal observations. However, due to the frequent and uncontrolled changes in the observation viewpoint, existing handheld systems have not considered point cloud–based data collection frameworks.

\begin{figure*}[h]
	\centering
	\includegraphics[width=1\linewidth]{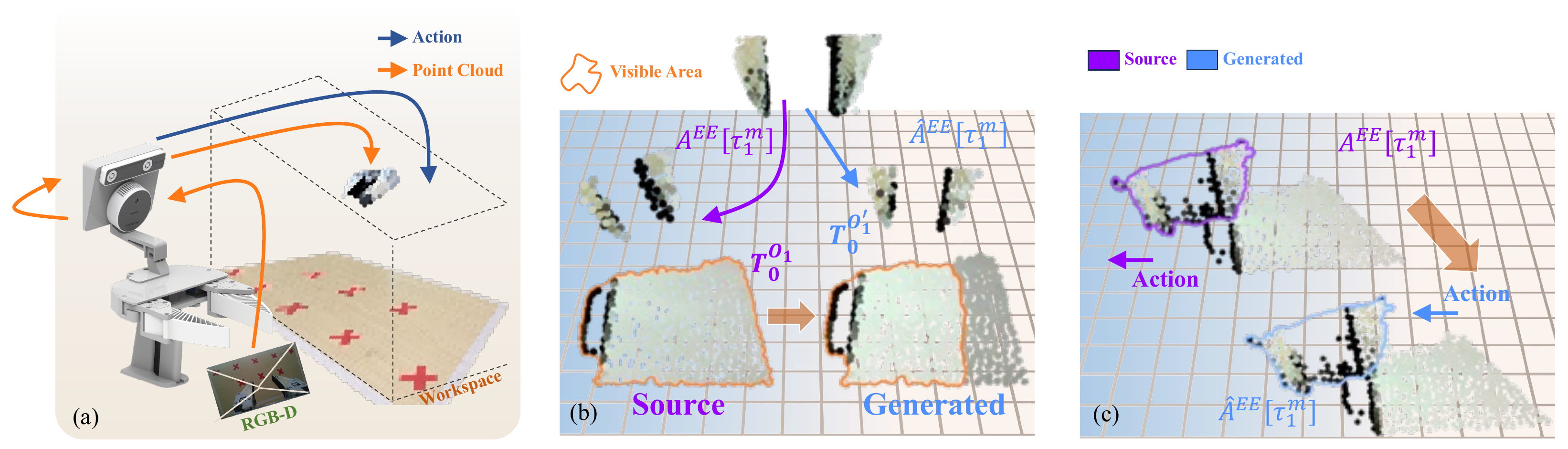}
	\caption{Overview of the dataset collection and generation pipeline. (a) The collection of observation–action pairs, where orange arrows transform the point cloud from the camera coordinate to the robot base coordinate, while blue arrows mark the corresponding 6D action pose. (b) The motion stage plans actions that bridge adjacent manipulation segments. The point cloud is cropped to the camera visible region and used as the generated observation. (c) The manipulation stage applies a transformation to all actions.} \label{pipeline}
\end{figure*}

\subsection*{b) 3D Imitation Learning for Robotics}

Image-based imitation learning has advanced rapidly \cite{wang2023mimicplay,prasad2024consistency,wang2024equivariant}, yet it generalizes poorly to complex, unstructured scenes. Many works therefore move beyond images alone.

Policy base on 3D observations methods such as PerAct \cite{shridhar2023perceiver}, GNFactor \cite{ze2023gnfactor}, and ACT3D \cite{gervet2023act3d} deliver strong manipulation results in low-dimensional control. The 3D Diffusion Policy further excels across varied tasks, highlighting 3D observations as a powerful foundation for imitation learning.

IDP3 \cite{ze2024generalizable} pushes this line forward by using egocentric 3D inputs, which reduce the need for precise camera calibration and fine segmentation and extend 3D policies to more realistic settings.

\subsection*{c) Data Generation for Robotic Manipulation}




Automatic demonstration generation reduces the need for manual data collection and powers modern imitation learning. MimicGen family~\cite{mandlekar2023mimicgen,jiang2024dexmimicgen,garrett2024skillmimicgen} adapts human demonstrations to new object setups but still relies on time-intensive real robot rollouts. 
DemoGen~\cite{xue2025demogen} eliminates this bottleneck by using a fully synthetic point cloud segmentation pipeline that produces smooth, executable demonstrations at low cost.

 
\section{Method}

UMIGen begins with Cloud-UMI, a handheld data collection system built upon UMI, designed to naturally capture human demonstrations with point cloud observations. 

For demonstrations captured in the world frame, we extend DemoGen to operate with point cloud observations from wrist-mounted view. This enables efficient and natural data collection and deployment across a variety of embodiments. The complete dataset collection and generation pipeline is illustrated in Fig.~\ref{pipeline}.

\subsection*{Hardware Design: Cloud-UMI}



Inspired by recent handheld motion data collection systems~\cite{chi2024universal,wu2024fast}, we develop Cloud-UMI: a low-cost, modular, and robot-independent point cloud collection device. As illustrated in Fig.~\ref{head} (b), the system consists of multiple interchangeable modules, including a handheld grip, a trigger mechanism, and mounting interfaces for depth and tracking sensors.

Cloud-UMI supports two observation modes, depending on how the captured point cloud is spatially anchored:

\begin{itemize}
    \item The system uses the tracking sensor to estimate the camera's pose and transform the captured point cloud into robot base frame.
    
    \item The raw point cloud is retained in the depth camera’s local coordinate frame, without transformation. This mode reflects purely egocentric observations and can be used directly for learning policies like IDP3 conditioned on point cloud in depth camera’s local coordinate frame.
\end{itemize}

\begin{figure*}[h]
	\centering
	\includegraphics[width=1\linewidth]{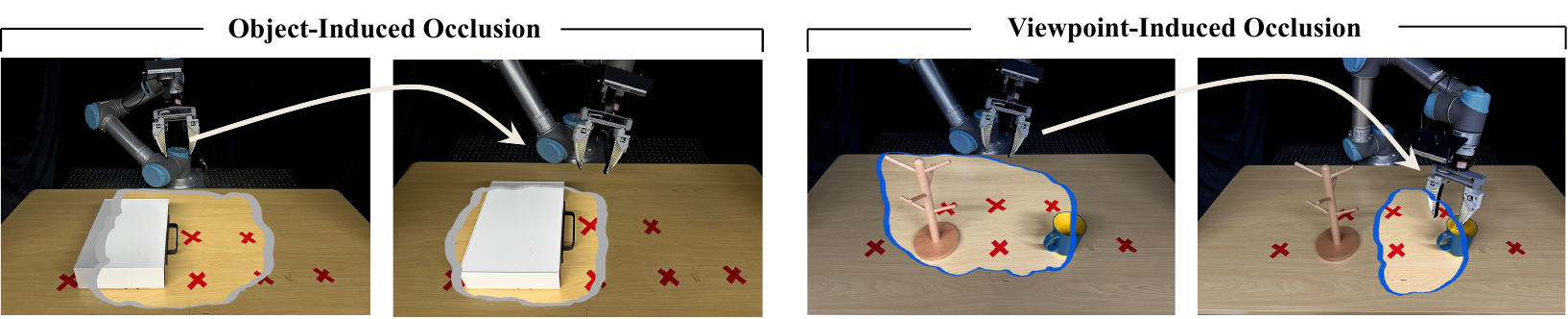}
      \caption{
      Illustration of two types of egocentric occlusions encountered during data collection. (Left) Object-Induced Occlusion: Large objects obstruct the camera's FoV, preventing visibility of surrounding workspace regions. (Right) Viewpoint-Induced Occlusion: The limited and task-dependent viewpoint of a wrist-mounted camera causes certain key elements of the task to fall outside the view during different stages of execution.
      } 
  \label{Occlusion}
\end{figure*}

This dual-mode design offers flexibility for various downstream applications, enabling both globally aligned and camera-centric representations depending on the task setting and embodiment constraints.

\subsubsection*{Depth Camera Module}

In prior UMI-related work, most systems employed fisheye cameras to capture wide-view images. Compared to conventional cameras, fisheye lenses offer a significantly larger FoV, providing broader observational coverage. However, in the context of point cloud based data collection, fisheye cameras are unable to produce high accuracy depth images. We selected the Intel RealSense L515 as our primary observation sensor. To mitigate issues arising from its narrower FoV, particularly occlusions and depth artifacts during manipulation, we made an adjustment to the sensor's mounting position. Specifically, we shifted the L515 slightly backwar,
an adjustment that simultaneously enlarges the visible workspace and reduces depth-loss problems when objects come too close to the sensor.

\subsubsection*{Pose Tracking Module}

To avoid complex calibration procedures, we adopt the Intel RealSense T265 for robust pose tracking, following the design choice in FastUMI and replacing the original visual odometry module used in UMI. While the T265 offers robust tracking capabilities and ease of deployment, its built-in IMU exhibits drift accumulation during extended use, making it less ideal for long term data collection scenarios. Interestingly, this limitation aligns well with our use of point cloud based observations: visuomotor policies learned from point cloud observations typically exhibit high data efficiency, requiring far fewer demonstrations compared to image-based alternatives.

\begin{figure*}[h]
	\centering
	\includegraphics[width=1\linewidth]{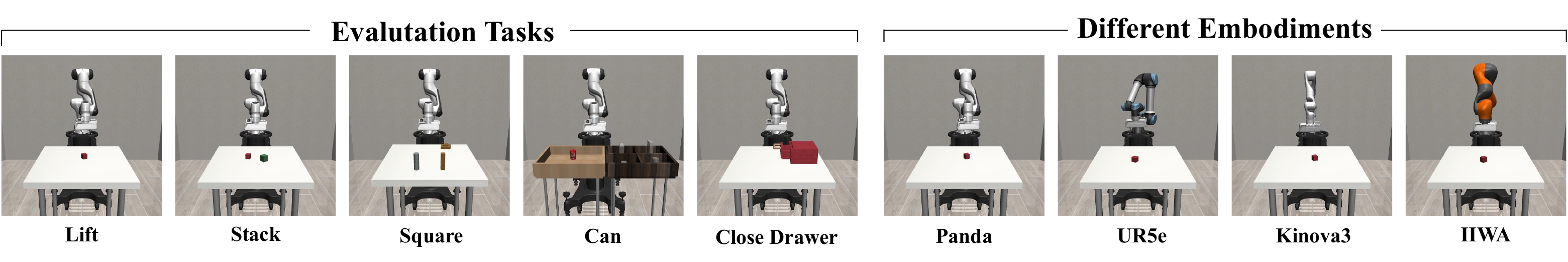}
      \caption{The simulation benchmark comprises five tasks (Lift, Stack, Square, Can, Close Drawer) and four robot arms (Panda, UR5e, Kinova 3, IIWA).
      } 
  \label{Sim}
\end{figure*}

\subsubsection*{Soft Gripper Module}

Traditional rigid grippers often suffer from slippage when handling objects with irregular or uneven surfaces, especially during real-world data collection. To improve grasp stability, we redesigned the gripper using flexible materials. The resulting soft gripper can partially conform to the shape of the target object during manipulation, enabling more secure and robust grasps across a wider range of object geometries.

At each timestep $t$, we capture an RGB-D image from a wrist-mounted depth camera and reconstruct a point cloud $P_t^{\text{cam}}$ in the camera's local coordinate frame.

To express this point cloud in the robot base frame (the fixed coordinate frame attached to the robot, serving as its spatial reference), we apply a calibrated transformation chain that involves two reference frames: the pose estimation frame and the robot base frame. The pose estimation system provides the 6-DoF pose of the camera relative to its own initial frame, which we denote as $T_{t}^{\text{pose}} \in \mathrm{SE}(3)$. This frame serves as a local world frame for accumulating motion over time.

We first compute the point cloud in the pose estimation frame as:
\begin{equation}
P_t^{\text{pose-initial}} = T_{t}^{\text{pose}}
    \, T_{\text{pose}\leftarrow\text{cam}}
    \, P_t^{\text{cam}},
\label{eq:pose_initial}
\end{equation}
where $T_{\text{pose} \leftarrow \text{cam}}$ is the fixed extrinsic calibration from the depth camera to the pose estimation device.

To convert this into the global robot base frame, we apply an additional rigid transformation $T_{\text{robot} \leftarrow \text{pose-initial}}$, which is obtained via offline calibration between the pose estimation origin and the robot base. The final point cloud in the robot frame is then:
\begin{equation}
P_t^{\text{robot}} = T_{\text{robot} \leftarrow \text{pose-initial}} \cdot P_t^{\text{pose-initial}}.
\label{eq:robot}
\end{equation}


Equivalently, we can compute the full 6-DoF pose of the tracking camera in the robot base frame as:
\begin{equation}
T_{t}^{\text{robot}} = T_{\text{robot} \leftarrow \text{pose-initial}} \cdot T_{t}^{\text{pose}}.
\label{eq:pose}
\end{equation}

Importantly, since the camera is rigidly attached to the end-effector, this pose also implicitly defines the robot's end-effector pose at time $t$, and can therefore serve directly as the action representation for control and imitation learning. By treating $T_{t}^{\text{robot}}$ as the target pose, we enable the reproduction of demonstrated trajectories using standard inverse kinematics or low-level motion controllers.

\subsection*{Synthetic Demonstration Generation}
DemoGen enables the generation of spatially augmented observation-action pairs from a small set of source demonstrations. In our work, we further extend DemoGen to handle wrist-mounted point cloud observations with VAO. We first revisit the preliminaries introduced in the original DemoGen framework.

\subsubsection*{Preliminaries}




As shown in Fig.~\ref{pipeline}, we consider the problem of visuomotor policy learning where a policy
$\pi : \mathcal{O} \rightarrow \mathcal{A}$ maps visual observations
$o \in \mathcal{O}$ to actions $a \in \mathcal{A}$.
Given a source demonstration
\begin{equation}
\mathcal{D}_{s_0} = \{(o_t, a_t)\}_{t=0}^{L-1},
\label{eq:demo}
\end{equation}
conditioned on an initial object configuration $s_0$, the goal of demonstration generation is to
create a new demonstration $\hat{\mathcal{D}}_{s_0'}$ under a novel configuration $s_0'$.

Each object configuration is defined by a set of $\mathrm{SE}(3)$ poses:
\begin{equation}
s_0 = \{T^O_0\}_{O=1}^{K}, 
\qquad
s_0' = \{T^{\prime O}_0\}_{O=1}^{K},
\label{eq:init_config}
\end{equation}
and the transformation between them is
\begin{equation}
\Delta s_0 \;=\; \bigl\{ \,(T^O_0)^{-1} T^{\prime O}_0 \,\bigr\}_{O=1}^{K}.
\label{eq:delta_config}
\end{equation}

The action at each time step includes both arm and hand components:
\begin{equation}
a_t = \bigl(a_t^{\text{arm}},\, a_t^{\text{hand}}\bigr),
\label{eq:action}
\end{equation}
where the end-effector poses $a_t^{\text{arm}} \in \mathrm{SE}(3)$ are spatially adapted based on
$\Delta s_0$, while the hand commands $a_t^{\text{hand}}$ remain invariant.

To ensure physical plausibility, the source trajectory is segmented into skill segments (contact-rich) and motion segments. Skill segments are transformed using the corresponding object's SE(3) transformation, and motion segments are replanned to connect adjacent skill segments. Corresponding observations are synthesized by applying the same transformations to segmented point clouds and proprioceptive states, yielding spatially consistent observation-action trajectories suitable for policy training.


\subsubsection*{Visibility-Aware Optimization (VAO)}

While traditional DemoGen assumes global visibility from fixed external viewpoints, wrist-mounted egocentric observations often suffer from limited FoV due to the dynamic pose of the end-effector. As illustrated in Fig.~\ref{Occlusion}, we define such visibility limitations as occlusions and further categorize them into two distinct types. 

The first type, which we refer to as \textbf{Object-Induced Occlusion}, occurs when large or bulky objects obstruct the egocentric camera's line of sight, leading to missing observations of nearby regions. The second type, termed \textbf{Viewpoint-Induced Occlusion}, arises in multi-stage or articulated tasks, where the camera's limited FoV caused by its wrist-mounted configuration fails to capture all relevant task elements during motion. 

Both types of occlusion introduce fundamental challenges for generating coherent demonstrations using global spatial transformations, as employed in DemoGen, and can result in unrealistic or inconsistent synthetic data due to partial scene understanding. To address the limited and viewpoint-dependent visibility of wrist-mounted cameras, we introduce VAO  that constrains each transformed point cloud to align with the camera's instantaneous FoV.




Let $\hat{P}_t$ denote the transformed point cloud at timestep $t$ generated by the generation pipeline.
To determine whether a point $p \in \hat{P}_t$ is visible from the camera at time $t$, we project it
onto the image plane using the known camera intrinsics $K$ and camera pose
$T_t^{\text{cam}}\!\in\!\mathrm{SE}(3)$:
\begin{equation}
\mathbf{u}
  \;=\;
  \Pi\!\Bigl( K \,\bigl(T_t^{\text{cam}}\bigr)^{-1} p \Bigr),
\label{eq:proj_u}
\end{equation}
where $\Pi(\cdot)$ denotes the perspective-projection operator that maps a 3D point to pixel
coordinates $\mathbf{u}=(u,v)$.  
We then define a binary visibility mask by checking whether $\mathbf{u}$ lies within the image bounds;
points outside the frame are treated as occluded or out-of-view and are discarded.  
The remaining points are re-projected back into the base frame via $T_t^{\text{cam}}$ to form the
filtered visible point cloud:
\begin{equation}
\hat{P}_t^{\text{visible}}
  \;=\;
  \Bigl\{\, p \in \hat{P}_t \;\Bigm|\; u \!\in\! [0,W),\; v \!\in\! [0,H) \Bigr\},
\label{eq:visible_pc}
\end{equation}
where $W$ and $H$ denote the image width and height, respectively.

To ensure a consistent input size for downstream learning, we apply farthest-point sampling (FPS) to
obtain the final point cloud:
\begin{equation}
\hat{P}_t^{\text{final}}
  \;=\;
  \operatorname{FPS}\!\bigl(\hat{P}_t^{\text{visible}},\, N\bigr).
\label{eq:fps}
\end{equation}

This process enforces egocentric visibility constraints on synthetic demonstrations, bridging the gap between idealized global observations and realistic wrist-mounted views.

\section{Experiments}

\begin{figure*}[h]
	\centering
	\includegraphics[width=1\linewidth]{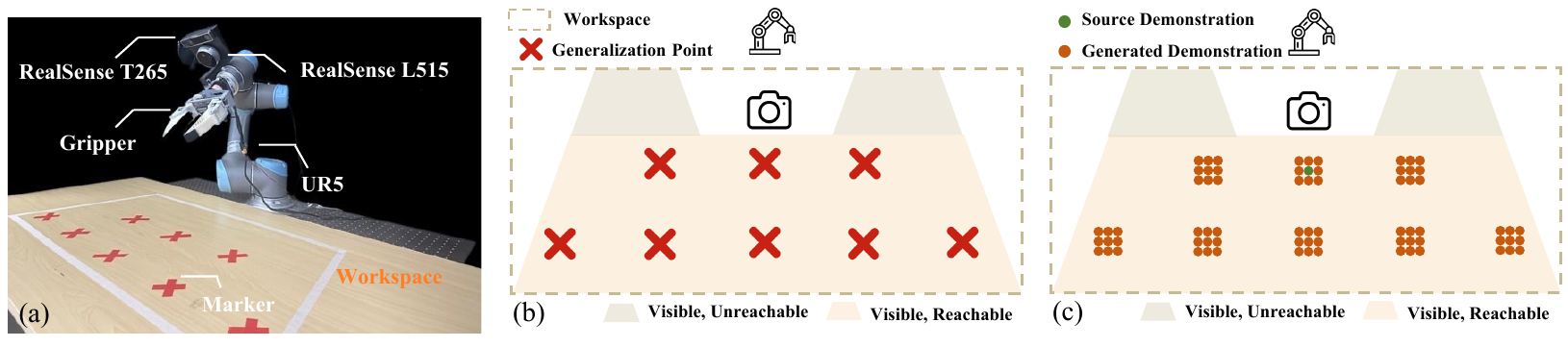}
	\caption{ Overview of the experimental setup and spatial generalization configuration.  
(a) Real-world hardware platform used for experiment tasks.  
(b) Workspace layout for spatial generalization evaluation. Red markers denote the generalization locations used to evaluate generalization performance.  
(c) Visualization of demonstration generation configuration. Green markers denote the location of source demonstrations, orange markers indicate candidate locations of generated demonstrations.} \label{system}
\end{figure*}

\begin{figure*}[h]
	\centering
	\includegraphics[width=1\linewidth]{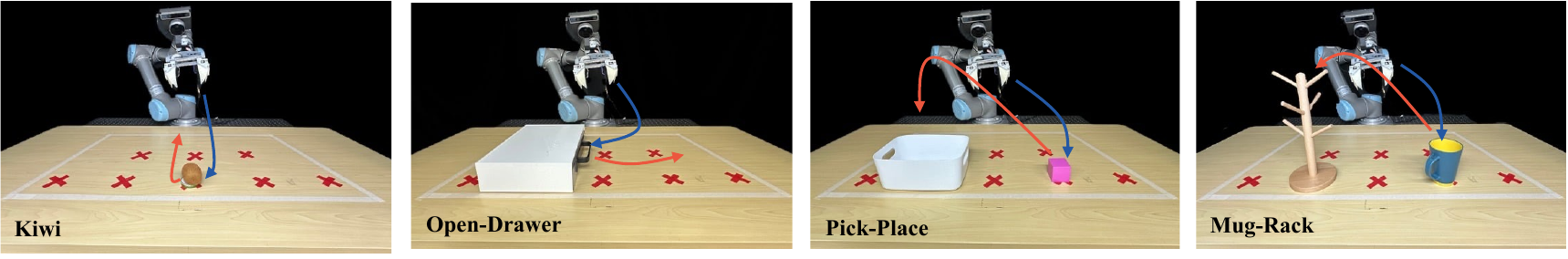}
	\caption{Spatial generalization tasks. Kiwi and Open-Drawer are single-stage tasks characterized by contact-intensive interactions, while Pick-Place and Mug-Rack are multi-stage tasks that demand precise sequential manipulation. Curved arrows trace the end-effector trajectory, blue segments indicate the gripper is open and orange segments indicate it is closed.} \label{system}
\end{figure*}

\subsection{Simulation Experiments}

Compared to observations in robot base coordinate, wrist-mounted perspectives generally provide more limited and task-dependent visual coverage due to the restricted FoV and occlusions introduced by the end-effector motion. To assess whether such constrained observations can still support high policy performance under equal data efficiency, we conduct a comparative study across five simulated manipulation tasks. To validate this hypothesis, we design a set of experiments as illustrated in Fig.~\ref{Sim}, incorporating multiple robot embodiments to further evaluate the consistency and generalization of the learned policies under diverse kinematic structures.

Since task complexity varies significantly across scenarios, we adjust the number of training demonstrations per task to ensure sufficient stability in evaluation. Specifically, we use 50 demonstrations for Lift and Close Drawer, 80 for Stack, and 150 for the more challenging tasks Can and Square.

For each task, we evaluate three configurations:
\begin{itemize}
    \item \textbf{Global-View DP3:} Policies trained using DP3 with frontview point cloud observations transformed to robot base coordinate.
    \item \textbf{Wrist-View DP3:} Policies trained using DP3 on wrist-mounted point clouds transformed to robot base coordinate.
    \item \textbf{Wrist-View IDP3:} Policies trained using IDP3, which is specifically designed for egocentric point cloud inputs in camera coordinate.
\end{itemize}



\begin{table}[h]
\centering
\begin{tabular}{lccc}
\toprule
\textbf{Task} & \textbf{W-DP3} & \textbf{G-DP3} & \textbf{W-IDP3} \\
\midrule
\textbf{Lift}          & 100.0\% & 100.0\% & 100.0\% \\
\textbf{Close Drawer}   & 100.0\% & 100.0\% & 93.3\% \\
\textbf{Can}            & 93.3\%  & 90.0\%  & 86.7\% \\
\textbf{Stack}          & 86.7\%  & 80.0\%  & 83.3\%  \\
\textbf{Square}         & 66.7\%  & 73.3\%  & 66.7\%  \\
\bottomrule
\end{tabular}
\caption{
Success rates on five simulated tasks under different observation-policy configurations.
W-DP3: DP3 trained with wrist-mounted observations;
G-DP3: DP3 trained with global observations from frontview;
W-IDP3: IDP3 trained with wrist-mounted observations.
}
\label{tab:wrist_global_comparison}
\end{table}

The results in Table~\ref{tab:wrist_global_comparison} support two key observations. First, although wrist-mounted observations inherently provide a narrower FoV, DP3 trained on these egocentric inputs still achieves success rates comparable to those using global observations, underscoring the data efficiency of point cloud–based policies even under partial observability. Second, despite operating directly in the camera coordinate frame without transforming point clouds to the robot base coordinate, the IDP3 variant still maintains competitive performance. This suggests that explicit global transformation is not strictly necessary for effective visuomotor policy learning in wrist-mounted settings.

However, point cloud–based models are inherently sensitive to outliers, particularly those caused by spurious or distant depth readings. In practice, we observe that egocentric perspectives are prone to capturing such noisy points, especially when reflective surfaces or cluttered scenes are present. To address this, our implementation of IDP3 includes a depth filtering step that removes points beyond a predefined distance threshold, effectively mitigating the impact of outliers and ensuring more stable policy learning.

To further assess the generalization capability of our method across different robot embodiments, we conduct a set of cross-embodiment experiments using four robot arms with distinct kinematic structures: Franka Panda, UR5e, Kinova Gen3, and KUKA IIWA.

For this evaluation, we train the DP3 algorithm using wrist-mounted observations collected on the Panda robot and deploy the resulting policy directly—without any finetuning—onto other robot embodiments. Success counts (out of 30 trials) for each configuration are reported in Table~\ref{tab:cross_embodiment}.

\begin{table}[h]
\centering
\begin{tabular}{lcccc}
\toprule
\textbf{Task} & \textbf{Panda} & \textbf{UR5e} & \textbf{Kinova3} & \textbf{IIWA} \\
\midrule
\textbf{Lift}          & 30/30 & 21/30 & 28/30 & 30/30 \\
\textbf{Close Drawer}  & 30/30 & 24/30 & 24/30 & 29/30 \\
\textbf{Can}      & 27/30 & 17/30 & 25/30 & 24/30 \\
\textbf{Stack}       & 25/30 & 16/30 & 14/30 &22/30 \\
\textbf{Square}       & 21/30 & 13/30 & 17/30 & 19/30 \\
\bottomrule
\end{tabular}
\caption{Cross-embodiment evaluation: success counts on Panda and other arms (out of 30 trials). }
\label{tab:cross_embodiment}
\end{table}

The results demonstrate that while variations in initial viewpoints and motion ranges across different robot arms may lead to differences in absolute success rates, the policy remains transferable across embodiments. Despite these embodiment-specific discrepancies, the consistent success patterns indicate that wrist-mounted point cloud observations and the proposed policy framework possess strong cross-embodiment generalization capability.

We assess the spatial generalization capability across 4 real-world tasks. A task summary is provided in Table~\ref{tab:spatial_tasks_simple}.

\begin{table}[t]
  \centering
  \setlength{\tabcolsep}{1mm}  

  \begin{tabular}{lccccc}
    \toprule
    \textbf{Task} & \textbf{\#Obj} & \textbf{\#SDemo} & \textbf{\#GDemo} & \textbf{\#Eval} & \textbf{\#Occ} \\
    \midrule
    \textbf{Kiwi}        & 1 & 3 & 3~$\times$~9~$\times$~8 & 8 & \xmark   \\
    \textbf{Open-Drawer} & 1 & 3 & 3~$\times$~9~$\times$~5 & 5 & \cmark   \\
    \textbf{Mug-Rack}    & 2 & 6 & 6~$\times$~9~$\times$~5 & 5 & \cmark   \\
    \textbf{Pick-Place}  & 2 & 6 & 6~$\times$~9~$\times$~5 & 5 & \cmark   \\
    \bottomrule
  \end{tabular}

  \caption{Real-world spatial-generalization tasks. \#Obj: number of manipulated objects; \#SDemo: human-collected source demonstrations; \#GDemo: generated demonstrations; \#Eval: evaluated configurations; \#Occ: presence of occlusion.}
  \label{tab:spatial_tasks_simple}
\end{table}

All experiments are conducted on a UR5 single-arm platform equipped with a wrist-mounted depth camera. To evaluate generalization under realistic spatial variations, we follow the protocol established in DemoGen. For each evaluated configuration, we introduce random spatial perturbations of $(\pm1.5~\text{cm}) \times (\pm1.5~\text{cm})$ around the initial object positions to generate 9 demonstration samples. This setup mimics common placement variations encountered in real-world scenarios. In summary, the total number of generated demonstrations per task is computed as $(\#\text{SDemo}) \times (\#\text{Eval}) \times 9$.


In real-world experiments, point cloud observations are often noisy and incomplete, with artifacts such as flickering holes, discontinuities, and distortions near object boundaries. Training on limited or homogeneous data under such partial observability can cause overfitting. Unlike DemoGen, which augments diversity by replaying the same trajectory , our data collection system benefits from its portability. We directly use the handheld device to capture multiple semantically similar trajectories. This introduces natural variation and avoids overly repetitive observations, improving policy robustness and generalization.

\begin{figure*}[h]
	\centering
	\includegraphics[width=1\linewidth]{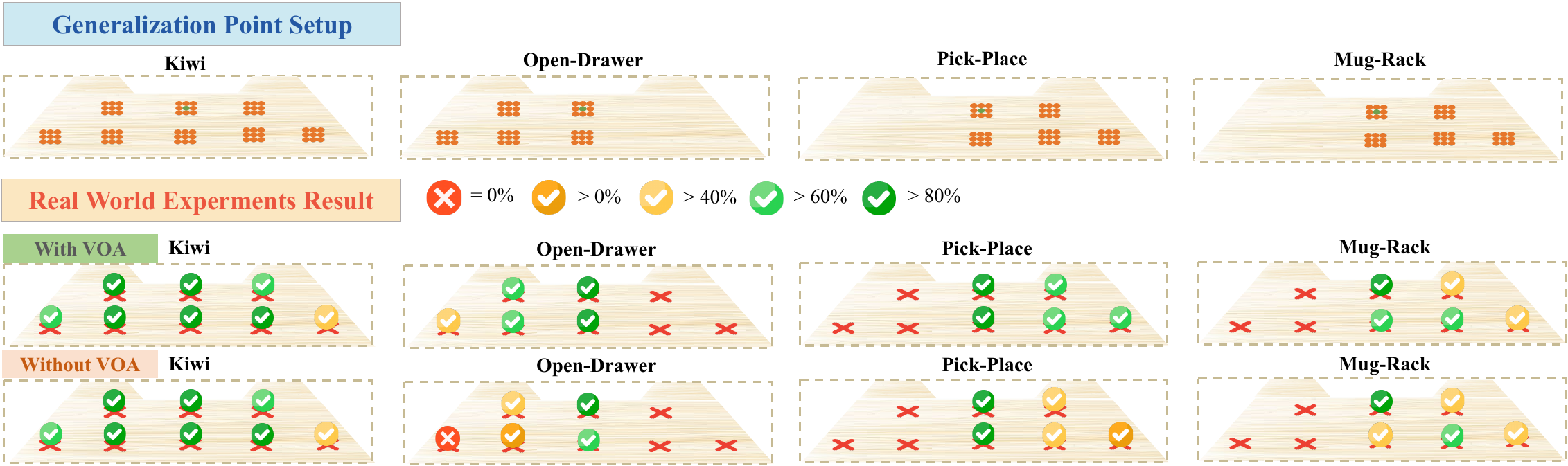}
	\caption{Generalization points and success rate heatmaps for all real-world evaluation tasks. For multi-stage tasks, each coordinate represents a placement of the manipulated object, while the non-manipulated object remains fixed. The success rates with and without VAO show how VAO affects performance across different spatial configurations.} \label{generalization}
\end{figure*}

Fig.~\ref{generalization} shows generalization points of different tasks and success rate heatmaps for all evaluation tasks. In our experimental setup, special adjustments are made for scenarios involving occlusions, as illustrated in Fig.~\ref{Occlusion}. For tasks without occlusions, we evaluate all generalization points within the defined workspace. However, for occlusion tasks, we selectively generate demonstrations only from a subset of generalization points to avoid large perceptual mismatches between generated and real observations.

For multi-stage tasks such as Pick-Place and Mug-Rack, we apply spatial generalization only to the manipulated object while keeping the positions of the basket or rack fixed to ensure physical feasibility in real-world experiments. This strategy enables consistent and safe task execution in constrained environments. Thanks to the simplicity and efficiency of our data collection system, we are able to significantly accelerate the data collection process. 
\subsection{Real Experiments}

As shown in the heatmaps in Fig.~\ref{generalization}, our approach demonstrates strong spatial generalization capabilities across a variety of tasks. Notably, we achieve high success rates (often exceeding 80\%) at most generalization points for single-stage tasks such as Kiwi and Open-Drawer. Even in more complex, multi-stage or occlusion-heavy tasks like Mug-Rack and Pick-Place , our system maintains robust performance, significantly outperforming baseline expectations. Furthermore, we observe consistent performance improvements when applying VAO: success rates increase across all generalization points compared to training without VAO, highlighting the critical role of VAO in producing realistic and physically grounded point cloud observations.

We observe that tasks performed near the center of the workspace tend to have higher success rates, while performance near workspace boundaries is less stable. We attribute this to two primary factors: (1) generalization points that are farther in Euclidean distance from the original demonstrations introduce larger differences in wrist-mounted observations, which may reduce data fidelity; (2) robot kinematic limits near workspace edges can prevent successful execution of intended trajectories.




\section*{Conclusion}

UMIGen provides a practical solution to the data bottleneck in robotic imitation learning by enabling fast, low-cost collection and generation of high-quality point cloud–action pairs.  Using the handheld Cloud-UMI device, data can be captured without visual SLAM or robot hardware.  UMIGen further extends the DemoGen framework with VAO that generates egocentric observations aligned with the camera’s actual FoV.  Empirical results show that VAO consistently improves success rates across generalization points, confirming its role in producing realistic and transferable training data.

Due to its hardware simplicity and high data efficiency, UMIGen dramatically reduces the cost and time of data acquisition. Moreover, the system naturally supports spatial generalization and cross-embodiment transfer without any additional post-processing.

However, the approach depends on precise calibration. Small shifts between the depth sensor and the end effector can add local noise and limit fast or dynamic motions. UMIGen also relies on depth cameras rather than fisheye optics, which restrict coverage in tasks that span a large workspace.


\bibliographystyle{plainnat}
\bibliography{ref}

@inproceedings{mandlekar2018roboturk,
  title={Roboturk: A crowdsourcing platform for robotic skill learning through imitation},
  author={Mandlekar, Ajay and Zhu, Yuke and Garg, Animesh and Booher, Jonathan and Spero, Max and Tung, Albert and Gao, Julian and Emmons, John and Gupta, Anchit and Orbay, Emre and others},
  booktitle={Conference on Robot Learning},
  pages={879--893},
  year={2018},
  organization={PMLR}
}

@article{chi2023diffusion,
  title={Diffusion policy: Visuomotor policy learning via action diffusion},
  author={Chi, Cheng and Xu, Zhenjia and Feng, Siyuan and Cousineau, Eric and Du, Yilun and Burchfiel, Benjamin and Tedrake, Russ and Song, Shuran},
  journal={The International Journal of Robotics Research},
  pages={02783649241273668},
  year={2023},
  publisher={SAGE Publications Sage UK: London, England}
}

@article{ze20243d,
  title={3d diffusion policy: Generalizable visuomotor policy learning via simple 3d representations},
  author={Ze, Yanjie and Zhang, Gu and Zhang, Kangning and Hu, Chenyuan and Wang, Muhan and Xu, Huazhe},
  journal={arXiv preprint arXiv:2403.03954},
  year={2024}
}

@article{chi2024universal,
  title={Universal manipulation interface: In-the-wild robot teaching without in-the-wild robots},
  author={Chi, Cheng and Xu, Zhenjia and Pan, Chuer and Cousineau, Eric and Burchfiel, Benjamin and Feng, Siyuan and Tedrake, Russ and Song, Shuran},
  journal={arXiv preprint arXiv:2402.10329},
  year={2024}
}

@article{fu2024mobile,
  title={Mobile aloha: Learning bimanual mobile manipulation with low-cost whole-body teleoperation},
  author={Fu, Zipeng and Zhao, Tony Z and Finn, Chelsea},
  journal={arXiv preprint arXiv:2401.02117},
  year={2024}
}

@article{lin2024data,
  title={Data scaling laws in imitation learning for robotic manipulation},
  author={Lin, Fanqi and Hu, Yingdong and Sheng, Pingyue and Wen, Chuan and You, Jiacheng and Gao, Yang},
  journal={arXiv preprint arXiv:2410.18647},
  year={2024}
}

@inproceedings{o2024open,
  title={Open x-embodiment: Robotic learning datasets and rt-x models: Open x-embodiment collaboration 0},
  author={O’Neill, Abby and Rehman, Abdul and Maddukuri, Abhiram and Gupta, Abhishek and Padalkar, Abhishek and Lee, Abraham and Pooley, Acorn and Gupta, Agrim and Mandlekar, Ajay and Jain, Ajinkya and others},
  booktitle={2024 IEEE International Conference on Robotics and Automation (ICRA)},
  pages={6892--6903},
  year={2024},
  organization={IEEE}
}

@inproceedings{shridhar2023perceiver,
  title={Perceiver-actor: A multi-task transformer for robotic manipulation},
  author={Shridhar, Mohit and Manuelli, Lucas and Fox, Dieter},
  booktitle={Conference on Robot Learning},
  pages={785--799},
  year={2023},
  organization={PMLR}
}

@inproceedings{ze2023gnfactor,
  title={Gnfactor: Multi-task real robot learning with generalizable neural feature fields},
  author={Ze, Yanjie and Yan, Ge and Wu, Yueh-Hua and Macaluso, Annabella and Ge, Yuying and Ye, Jianglong and Hansen, Nicklas and Li, Li Erran and Wang, Xiaolong},
  booktitle={Conference on robot learning},
  pages={284--301},
  year={2023},
  organization={PMLR}
}

@article{gervet2023act3d,
  title={Act3d: 3d feature field transformers for multi-task robotic manipulation},
  author={Gervet, Theophile and Xian, Zhou and Gkanatsios, Nikolaos and Fragkiadaki, Katerina},
  journal={arXiv preprint arXiv:2306.17817},
  year={2023}
}

@article{ze2024generalizable,
  title={Generalizable humanoid manipulation with 3d diffusion policies},
  author={Ze, Yanjie and Chen, Zixuan and Wang, Wenhao and Chen, Tianyi and He, Xialin and Yuan, Ying and Peng, Xue Bin and Wu, Jiajun},
  journal={arXiv preprint arXiv:2410.10803},
  year={2024}
}

@article{wu2024fast,
  title={Fast-umi: A scalable and hardware-independent universal manipulation interface},
  author={Wu, Ziniu and Wang, Tianyu and Guan, Chuyue and Jia, Zhongjie and Liang, Shuai and Song, Haoming and Qu, Delin and Wang, Dong and Wang, Zhigang and Cao, Nieqing and others},
  journal={arXiv e-prints},
  pages={arXiv--2409},
  year={2024}
}

@article{mandlekar2023mimicgen,
  title={Mimicgen: A data generation system for scalable robot learning using human demonstrations},
  author={Mandlekar, Ajay and Nasiriany, Soroush and Wen, Bowen and Akinola, Iretiayo and Narang, Yashraj and Fan, Linxi and Zhu, Yuke and Fox, Dieter},
  journal={arXiv preprint arXiv:2310.17596},
  year={2023}
}

@article{jiang2024dexmimicgen,
  title={Dexmimicgen: Automated data generation for bimanual dexterous manipulation via imitation learning},
  author={Jiang, Zhenyu and Xie, Yuqi and Lin, Kevin and Xu, Zhenjia and Wan, Weikang and Mandlekar, Ajay and Fan, Linxi and Zhu, Yuke},
  journal={arXiv preprint arXiv:2410.24185},
  year={2024}
}

@article{xue2025demogen,
  title={Demogen: Synthetic demonstration generation for data-efficient visuomotor policy learning},
  author={Xue, Zhengrong and Deng, Shuying and Chen, Zhenyang and Wang, Yixuan and Yuan, Zhecheng and Xu, Huazhe},
  journal={arXiv preprint arXiv:2502.16932},
  year={2025}
}

@article{zhao2024aloha,
  title={Aloha unleashed: A simple recipe for robot dexterity},
  author={Zhao, Tony Z and Tompson, Jonathan and Driess, Danny and Florence, Pete and Ghasemipour, Kamyar and Finn, Chelsea and Wahid, Ayzaan},
  journal={arXiv preprint arXiv:2410.13126},
  year={2024}
}

@article{garrett2024skillmimicgen,
  title={Skillmimicgen: Automated demonstration generation for efficient skill learning and deployment},
  author={Garrett, Caelan and Mandlekar, Ajay and Wen, Bowen and Fox, Dieter},
  journal={arXiv preprint arXiv:2410.18907},
  year={2024}
}

@article{yang2025fp3,
  title={Fp3: A 3d foundation policy for robotic manipulation},
  author={Yang, Rujia and Chen, Geng and Wen, Chuan and Gao, Yang},
  journal={arXiv preprint arXiv:2503.08950},
  year={2025}
}

@article{ke20243d,
  title={3d diffuser actor: Policy diffusion with 3d scene representations},
  author={Ke, Tsung-Wei and Gkanatsios, Nikolaos and Fragkiadaki, Katerina},
  journal={arXiv preprint arXiv:2402.10885},
  year={2024}
}

@misc{khazatsky2025droidlargescaleinthewildrobot,
      title={DROID: A Large-Scale In-The-Wild Robot Manipulation Dataset}, 
      author={Alexander Khazatsky and Karl Pertsch and Suraj Nair and Ashwin Balakrishna and Sudeep Dasari and Siddharth Karamcheti and Soroush Nasiriany and Mohan Kumar Srirama and Lawrence Yunliang Chen and Kirsty Ellis and Peter David Fagan and Joey Hejna and Masha Itkina and Marion Lepert and Yecheng Jason Ma and Patrick Tree Miller and Jimmy Wu and Suneel Belkhale and Shivin Dass and Huy Ha and Arhan Jain and Abraham Lee and Youngwoon Lee and Marius Memmel and Sungjae Park and Ilija Radosavovic and Kaiyuan Wang and Albert Zhan and Kevin Black and Cheng Chi and Kyle Beltran Hatch and Shan Lin and Jingpei Lu and Jean Mercat and Abdul Rehman and Pannag R Sanketi and Archit Sharma and Cody Simpson and Quan Vuong and Homer Rich Walke and Blake Wulfe and Ted Xiao and Jonathan Heewon Yang and Arefeh Yavary and Tony Z. Zhao and Christopher Agia and Rohan Baijal and Mateo Guaman Castro and Daphne Chen and Qiuyu Chen and Trinity Chung and Jaimyn Drake and Ethan Paul Foster and Jensen Gao and Vitor Guizilini and David Antonio Herrera and Minho Heo and Kyle Hsu and Jiaheng Hu and Muhammad Zubair Irshad and Donovon Jackson and Charlotte Le and Yunshuang Li and Kevin Lin and Roy Lin and Zehan Ma and Abhiram Maddukuri and Suvir Mirchandani and Daniel Morton and Tony Nguyen and Abigail O'Neill and Rosario Scalise and Derick Seale and Victor Son and Stephen Tian and Emi Tran and Andrew E. Wang and Yilin Wu and Annie Xie and Jingyun Yang and Patrick Yin and Yunchu Zhang and Osbert Bastani and Glen Berseth and Jeannette Bohg and Ken Goldberg and Abhinav Gupta and Abhishek Gupta and Dinesh Jayaraman and Joseph J Lim and Jitendra Malik and Roberto Martín-Martín and Subramanian Ramamoorthy and Dorsa Sadigh and Shuran Song and Jiajun Wu and Michael C. Yip and Yuke Zhu and Thomas Kollar and Sergey Levine and Chelsea Finn},
      year={2025},
      eprint={2403.12945},
      archivePrefix={arXiv},
      primaryClass={cs.RO},
      url={https://arxiv.org/abs/2403.12945}, 
}

@article{Campos_2021,
   title={ORB-SLAM3: An Accurate Open-Source Library for Visual, Visual–Inertial, and Multimap SLAM},
   volume={37},
   ISSN={1941-0468},
   url={http://dx.doi.org/10.1109/TRO.2021.3075644},
   DOI={10.1109/tro.2021.3075644},
   number={6},
   journal={IEEE Transactions on Robotics},
   publisher={Institute of Electrical and Electronics Engineers (IEEE)},
   author={Campos, Carlos and Elvira, Richard and Rodriguez, Juan J. Gomez and M. Montiel, Jose M. and D. Tardos, Juan},
   year={2021},
   month=dec, pages={1874–1890} }

@misc{zhao2023learningfinegrainedbimanualmanipulation,
      title={Learning Fine-Grained Bimanual Manipulation with Low-Cost Hardware}, 
      author={Tony Z. Zhao and Vikash Kumar and Sergey Levine and Chelsea Finn},
      year={2023},
      eprint={2304.13705},
      archivePrefix={arXiv},
      primaryClass={cs.RO},
      url={https://arxiv.org/abs/2304.13705}, 
}

@inproceedings{wu2024gello,
  title={Gello: A general, low-cost, and intuitive teleoperation framework for robot manipulators},
  author={Wu, Philipp and Shentu, Yide and Yi, Zhongke and Lin, Xingyu and Abbeel, Pieter},
  booktitle={2024 IEEE/RSJ International Conference on Intelligent Robots and Systems (IROS)},
  pages={12156--12163},
  year={2024},
  organization={IEEE}
}

@inproceedings{zhu2023viola,
  title={Viola: Imitation learning for vision-based manipulation with object proposal priors},
  author={Zhu, Yifeng and Joshi, Abhishek and Stone, Peter and Zhu, Yuke},
  booktitle={Conference on Robot Learning},
  pages={1199--1210},
  year={2023},
  organization={PMLR}
}

@inproceedings{seo2023deep,
  title={Deep imitation learning for humanoid loco-manipulation through human teleoperation},
  author={Seo, Mingyo and Han, Steve and Sim, Kyutae and Bang, Seung Hyeon and Gonzalez, Carlos and Sentis, Luis and Zhu, Yuke},
  booktitle={2023 IEEE-RAS 22nd International Conference on Humanoid Robots (Humanoids)},
  pages={1--8},
  year={2023},
  organization={IEEE}
}

@inproceedings{rosete2023latent,
  title={Latent plans for task-agnostic offline reinforcement learning},
  author={Rosete-Beas, Erick and Mees, Oier and Kalweit, Gabriel and Boedecker, Joschka and Burgard, Wolfram},
  booktitle={Conference on Robot Learning},
  pages={1838--1849},
  year={2023},
  organization={PMLR}
}

@inproceedings{toedtheide2023force,
  title={A force-sensitive exoskeleton for teleoperation: An application in elderly care robotics},
  author={Toedtheide, Alexander and Chen, Xiao and Sadeghian, Hamid and Naceri, Abdeldjallil and Haddadin, Sami},
  booktitle={2023 IEEE International Conference on Robotics and Automation (ICRA)},
  pages={12624--12630},
  year={2023},
  organization={IEEE}
}

@misc{liu2025factrforceattendingcurriculumtraining,
      title={FACTR: Force-Attending Curriculum Training for Contact-Rich Policy Learning}, 
      author={Jason Jingzhou Liu and Yulong Li and Kenneth Shaw and Tony Tao and Ruslan Salakhutdinov and Deepak Pathak},
      year={2025},
      eprint={2502.17432},
      archivePrefix={arXiv},
      primaryClass={cs.RO},
      url={https://arxiv.org/abs/2502.17432}, 
}

@misc{zhu2025touchwildlearningfinegrained,
      title={Touch in the Wild: Learning Fine-Grained Manipulation with a Portable Visuo-Tactile Gripper}, 
      author={Xinyue Zhu and Binghao Huang and Yunzhu Li},
      year={2025},
      eprint={2507.15062},
      archivePrefix={arXiv},
      primaryClass={cs.RO},
      url={https://arxiv.org/abs/2507.15062}, 
}

@article{wang2023mimicplay,
  title={Mimicplay: Long-horizon imitation learning by watching human play},
  author={Wang, Chen and Fan, Linxi and Sun, Jiankai and Zhang, Ruohan and Fei-Fei, Li and Xu, Danfei and Zhu, Yuke and Anandkumar, Anima},
  journal={arXiv preprint arXiv:2302.12422},
  year={2023}
}

@article{prasad2024consistency,
  title={Consistency policy: Accelerated visuomotor policies via consistency distillation},
  author={Prasad, Aaditya and Lin, Kevin and Wu, Jimmy and Zhou, Linqi and Bohg, Jeannette},
  journal={arXiv preprint arXiv:2405.07503},
  year={2024}
}

@article{wang2024equivariant,
  title={Equivariant diffusion policy},
  author={Wang, Dian and Hart, Stephen and Surovik, David and Kelestemur, Tarik and Huang, Haojie and Zhao, Haibo and Yeatman, Mark and Wang, Jiuguang and Walters, Robin and Platt, Robert},
  journal={arXiv preprint arXiv:2407.01812},
  year={2024}
}

\end{document}